\documentclass[letterpaper, 10 pt, conference]{ieeeconf}  

\usepackage{graphicx}
\usepackage{subfigure}
\usepackage{amsmath}
\usepackage{float}
\usepackage{booktabs}
\usepackage{cite}
\usepackage{comment}
\usepackage{microtype}
\usepackage{array}
\usepackage{hyperref}
\hypersetup{
    colorlinks=true,
    linkcolor=blue,
    filecolor=magenta,      
    urlcolor=blue,
    }
\IEEEoverridecommandlockouts                              

\title{\fontsize{15.7}{19}\selectfont \bf
Design and Control of a Novel Six-Degree-of-Freedom Hybrid Robotic Arm
}

\author{Yang Chen$^1{^{,2}}$, Zhonghua Miao$^2$, Yuanyue Ge$^1$, Sen lin$^1$, Liping Chen$^1$* and Ya Xiong$^1$* 
\thanks{This work was supported by the Haidian District Bureau of Agriculture and Rural Affairs, the BAAFS Innovation Ability Project (KJCX20240321, KJICX20240502), the BAAFS Talent Recruitment Program, and the NSFC Excellent Young Scientists Fund (overseas).}
\thanks{$^1$Yang Chen, Yuanyue Ge, Sen Lin, Liping Chen and Ya Xiong are with the Intelligent Equipment Research Center, Beijing Academy of Agriculture and Forestry Sciences, Beijing 100097, China.} %
\thanks{$^2$Yang Chen and Zhonghua Miao are with the School of Mechanical Electrical Engineering and Automation, Shanghai University, Shanghai 20044, China.}%
\thanks{*Liping Chen and Ya Xiong are the corresponding authors,
        {\tt\small chenlp@nercita.org.cn, yaxiong@nercita.org.cn}.}%
}

\begin{document}

\maketitle
\thispagestyle{empty}
\pagestyle{empty}

\begin{abstract}

Robotic arms are key components in fruit-harvesting robots. In agricultural settings, conventional serial or parallel robotic arms often fall short in meeting the demands for a large workspace, rapid movement, enhanced capability of obstacle avoidance and affordability. This study proposes a novel hybrid six-degree-of-freedom (DoF) robotic arm that combines the advantages of parallel and serial mechanisms. Inspired by yoga, we designed two sliders capable of moving independently along a single rail, acting as two feet. These sliders are interconnected with linkages and a meshed-gear set, allowing the parallel mechanism to lower itself and perform a split to pass under obstacles. This unique feature allows the arm to avoid obstacles such as pipes, tables and beams typically found in greenhouses. Integrated with serially mounted joints, the patented hybrid arm is able to maintain the end's pose even when it moves with a mobile platform, facilitating fruit picking with the optimal pose in dynamic conditions. Moreover, the hybrid arm's workspace is substantially larger, being almost three times the volume of UR3 serial arms and fourteen times that of the ABB IRB parallel arms. Experiments show that the repeatability errors are 0.017 mm, 0.03 mm and 0.109 mm for the two sliders and the arm's end, respectively, providing sufficient precision for agricultural robots.
\end{abstract}

\section{INTRODUCTION}

Advancements in precision agriculture and artificial intelligence, coupled with the continual rise in labor costs, have made agricultural robots essential tools for replacing human labor and enhancing harvest efficiency \cite{sinha2022recent}. Taking high-value crops such as strawberries, tomatoes, eggplants, and sweet peppers as examples \cite{tao2021review}, the current harvesting operations are heavily reliant on human labor \cite{ren2020agricultural}. 
\begin{figure}[htbp]
\centering
\includegraphics[scale=0.18]{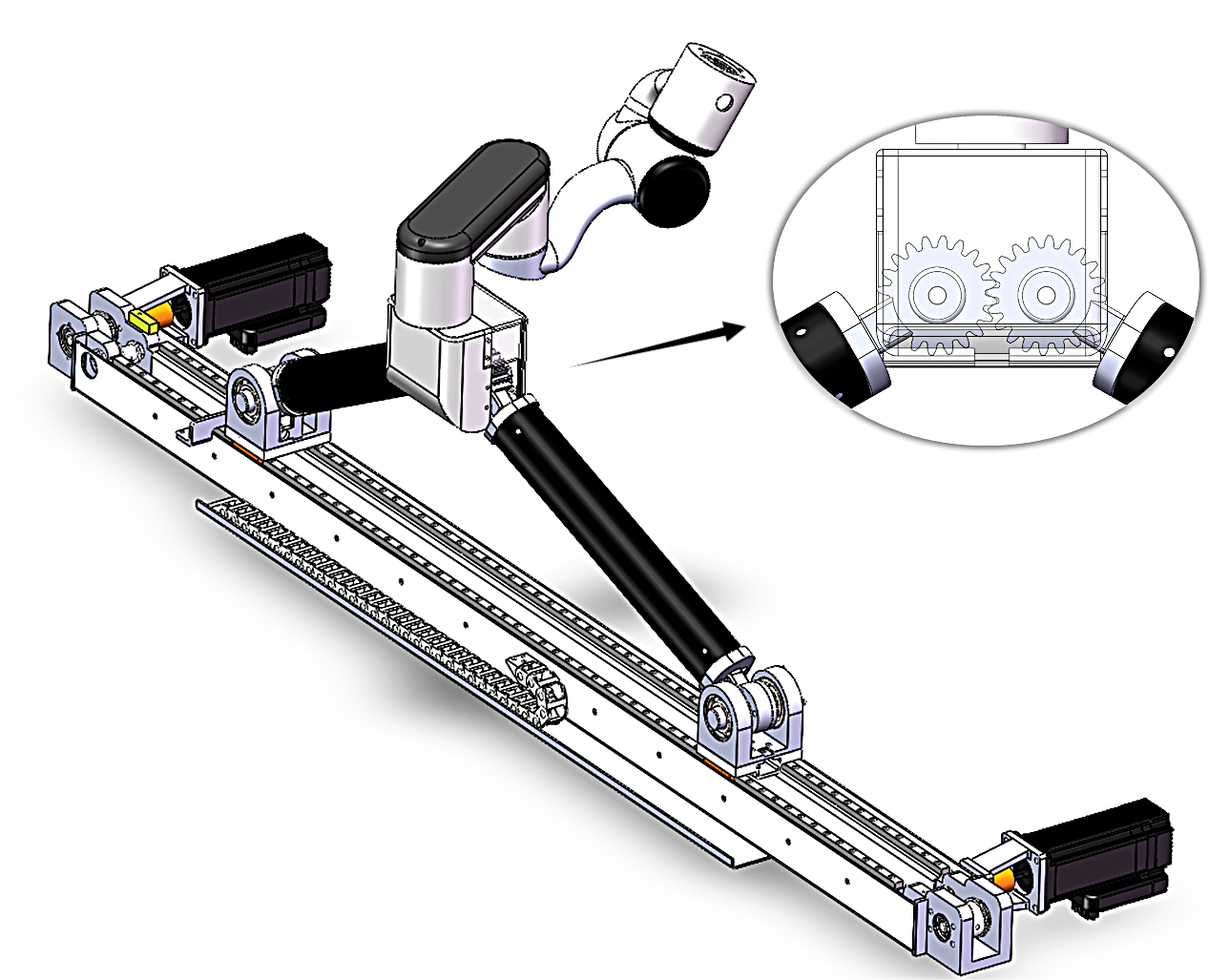}
\caption{The novel six-degree-of-freedom hybrid robotic arm; a demonstration video can be found at \url{https://drive.google.com/file/d/1vZgk7NcdaC_Dejxn6mOWbftSCUKCEhgD/view?usp=sharing}.}
\label{fig1}
\end{figure}

Autonomous harvesting robot system faces technical challenges in environmental adaptability \cite{xiong2018design}, fruit damage rate, harvesting speed, and power consumption \cite{xiong2020autonomous}. The robotic arm, as an actuator in harvesting robot systems, plays a crucial role in fruit harvesting and placement \cite{8594221}. However, a single configuration of the robotic arm is insufficient to meet the harvesting requirements of unstructured environments, especially for delicate and discretely distributed fruits and vegetables such as strawberries, necessitating the design of specialized robotic arms \cite{dong2022continuum}. Existing harvesting robotic arms are primarily available in three configurations: linear type, serial type, and parallel type. Linear arms, such as Cartesian robotic arms, feature simple structures \cite{barnett2020work}, good rigidity, easily controlled, and moderate speed but they require longer rails or joints during vertical movement, resulting in a lower spatial utilization rate and an increased overall size and space occupation \cite{huda2022implementation}. Serial robotic arms are expensive, complex and sensitive to slight deviations in target poses, rendering them unusable in scenarios requiring random fitting of target poses \cite{nguyen2020gravity}. Moreover, the complexity of the system reduces stability and motion speed \cite{rajpar2021reconfigurable}. Representatives of parallel arms, such as Delta robots, utilize a system of interconnected links and actuators arranged in parallel configurations \cite{heise2022effects}. Unlike serial arms, parallel arms operate with multiple degrees of freedom in parallel, providing high speed and high load-bearing capabilities, but their workspace is constrained by the parallel linkages forming its parallel structure \cite{hemming2020tutorial}. 
These arms occupy excessive space when not in operation, making it challenging for harvesting robots equipped with them to navigate through tight agricultural spaces such as pipes, tables and beams typically found in greenhouses.

Ensuring proficient trajectory tracking capability in robotic arms during harvesting is pivotal\cite{lavin2020trajectory}. Despite being equipped with advanced sensors and control systems enabling precise trajectory tracking and real-time adjustments, these robotic arms still face challenges in dynamically maintaining the end-effector's pose and passing through complex environments \cite{liu2021trajectory}. These challenges, influenced by factors such as mechanical design and control algorithms, may restrict their ability to achieve the desired precision and flexibility, particularly in tasks such as dynamically maintaining the optimal harvesting pose for a period \cite{al2023generalized}.


The primary contribution of this study is the proposal of a novel six-DoF hybrid robotic arm for fruit/vegetable harvesting, as depicted in Fig. \ref{fig1}. Its main novelties are outlined as follows:
\begin{itemize}

\item The arm was composed of a hybrid structure involving a serial mechanism and a single-rail dual-slider mechanism, featuring a large workspace and fast speed.
\item The hybrid arm incorporates a meshed-gear parallel mechanism, allowing it to occupy a small space when not in use and to duck under obstacles.
\item The robotic arm can remain the end-effector's pose unchanged for a while even when it moves with a mobile platform, making it possible for dynamic picking.
\end{itemize}

The rest of the article is organized as follows. Section II describes the design and manufacturing of the novel robotic arm. Section III introduces the forward and reverse kinematics analysis and control algorithms of the robotic arm. Section IV introduces some experiments, including functional demo, trajectory tracking test, and repetition test, which demonstrated the advantages and potential applications of the robotic arm.

\section{Design and manufacturing}

In the designing process, we aimed to reach several key goals: 1) to ensure the robotic arm occupies minimal space when not in use but expands to a larger operational workspace when in operation; 2) to maintain the end-effector's pose even when it moves with a mobile platform; 3) to achieve straightforward control; 4) to attain fast motion and precise positioning; 5) to have a design that is both compact and robust, combining strength with lightness.

Based on the requirements mentioned above, we designed the meshed-gear parallel mechanism and the single-rail dual-slider mechanism, integrated with several serially mounted revolute joints and finally obtained a patented hybrid robotic arm, as shown in Fig. \ref{fig1}.

\subsection{The meshed-gear parallel mechanism}

To make the end-effector's pose unchanged when the arm moves with a mobile platform, we came up with the idea of having a rail that allows the arm to travel in the opposite direction. Simply mounting an existing arm on the rail might make the arm difficult to use in compact environments, such as greenhouses. 

\begin{figure}[htbp]
\centering
\includegraphics[scale=0.21]{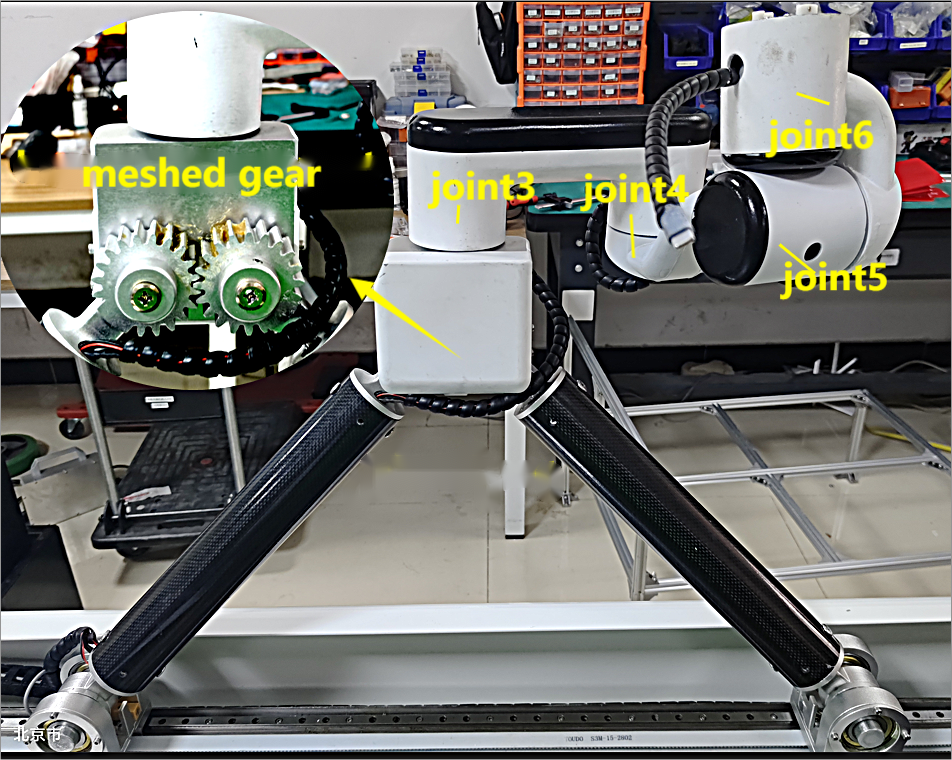}
\caption{The meshed-gear parallel mechanism assembly.}
\label{fig3}
\end{figure}

Inspired by yoga, we designed two sliders that can independently move along a rail, as shown in Fig. \ref{fig3}. The two sliders acted as feet. Connected by linkages, the upper part of the mechanism can lower itself to perform a split when the two sliders move in opposite directions. Conversely, the upper part can be raised when the two sliders move towards each other. This unique feature enabled the arm pass under obstacles. To keep the upper part parallel to the rail, we designed a meshed-gear mechanism. 
The gear ratio was set at 1: 1, which ensured that the angle of inclination between the upper serial part (joints 3-6) and the $x-y$ plane remains unchanged. This feature restricted the parallel mechanism to maintain parallel motion, serving as a replacement for the current parallel motion linkage mechanism. Consequently, it improved the stability of the end-effector and resulted in a more compact structure.

\subsection{The single-rail dual-slider mechanism}

\begin{figure}[htpb]
\centering
    \subfigure[]
{
    \begin{minipage}{.8\linewidth}
        \centering
    \includegraphics[scale=0.19]{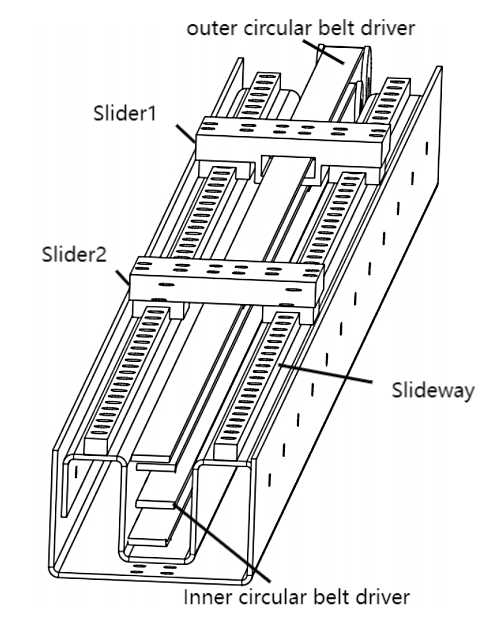}
    \end{minipage}
}
    \subfigure[]
{
    \begin{minipage}{.9\linewidth}
    \centering
    \includegraphics[scale=0.15]{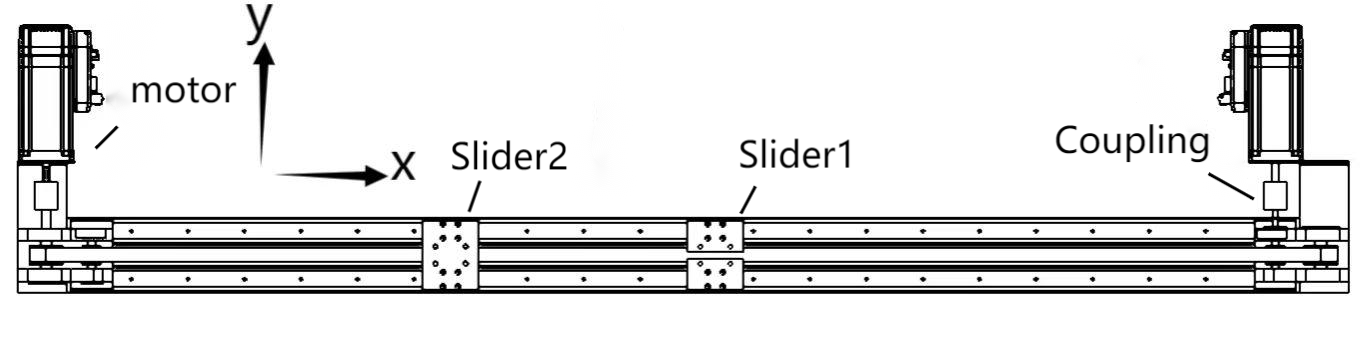}
    \end{minipage}
}
      \caption{Newly designed single-rail dual-slider mechanism: a) schematic diagram of the arrangement of the inner and outer circular belts; b) top view of the single-rail dual-slider mechanism.}
      \label{fig4}
   \end{figure}

Existing single-rail dual-slider mechanism allows for the independent movement of both sliders. However, the parallel arrangement of the two slider transmission mechanisms takes up considerable axial space, making them unsuitable for installations in compact workspaces.
As illustrated in Fig. \ref{fig4}, our patented design placed one slider transmission mechanism insider of the other slider transmission mechanism, forming an inner-outer timing belt mechanism that reduced space needed for independent movement of the two sliders. This design achieved unique capability of having dual-end inputs and independently driven sliders. The independent motion of the two sliders allowed for shared loading, enabling free linear movement in the same or opposite directions on the guide rail. 
\begin{figure}[htbp]
\centering
\includegraphics[scale=0.11]{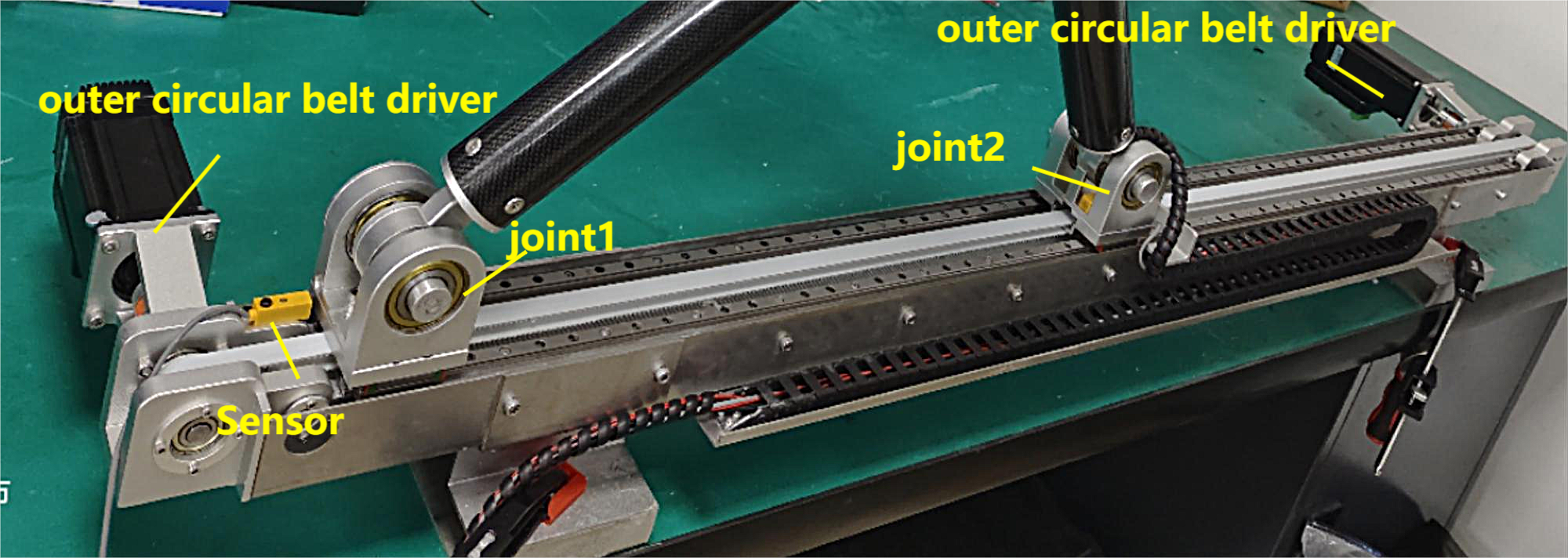}
\caption{Prototype of the patented single-rail dual-slider mechanism}
\label{fig5}
\end{figure}

 Fig. \ref{fig5} shows a prototype of the single-rail dual-slider mechanism, capable of achieving independent motion of the upper attached parts in multiple directions. It mainly consists of components such as timing belts, linear rails, aluminum profiles, bearings, motors, sensors, etc. The power input shafts were located on both sides of the linear module's driver shafts, with slider blocks fixed on the belt for installing upper attachments. 

\subsection{The hybrid robotic arm}

\begin{figure}[htbp]
\centering
\includegraphics[scale=0.34]{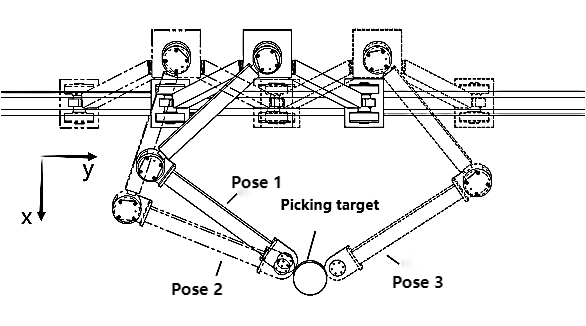}
\caption{ The meshed-gear parallel mechanism design.}
\label{fig2}
\end{figure}

By combining the proposed parallel mechanism with two serially mounted revolute joints, we first obtained a four-DoF hybrid arm for controlling the three-dimensional position and angles of the end effector. Fig. \ref{fig2} illustrates the operational principle of the proposed hybrid mechanism, allowing control the end-effector orientations around one picking target through coordinated motion between the bottom dual-slider mechanism and the upper rotating joints. 

\begin{figure}[htbp]
\centering
\includegraphics[scale=0.21]{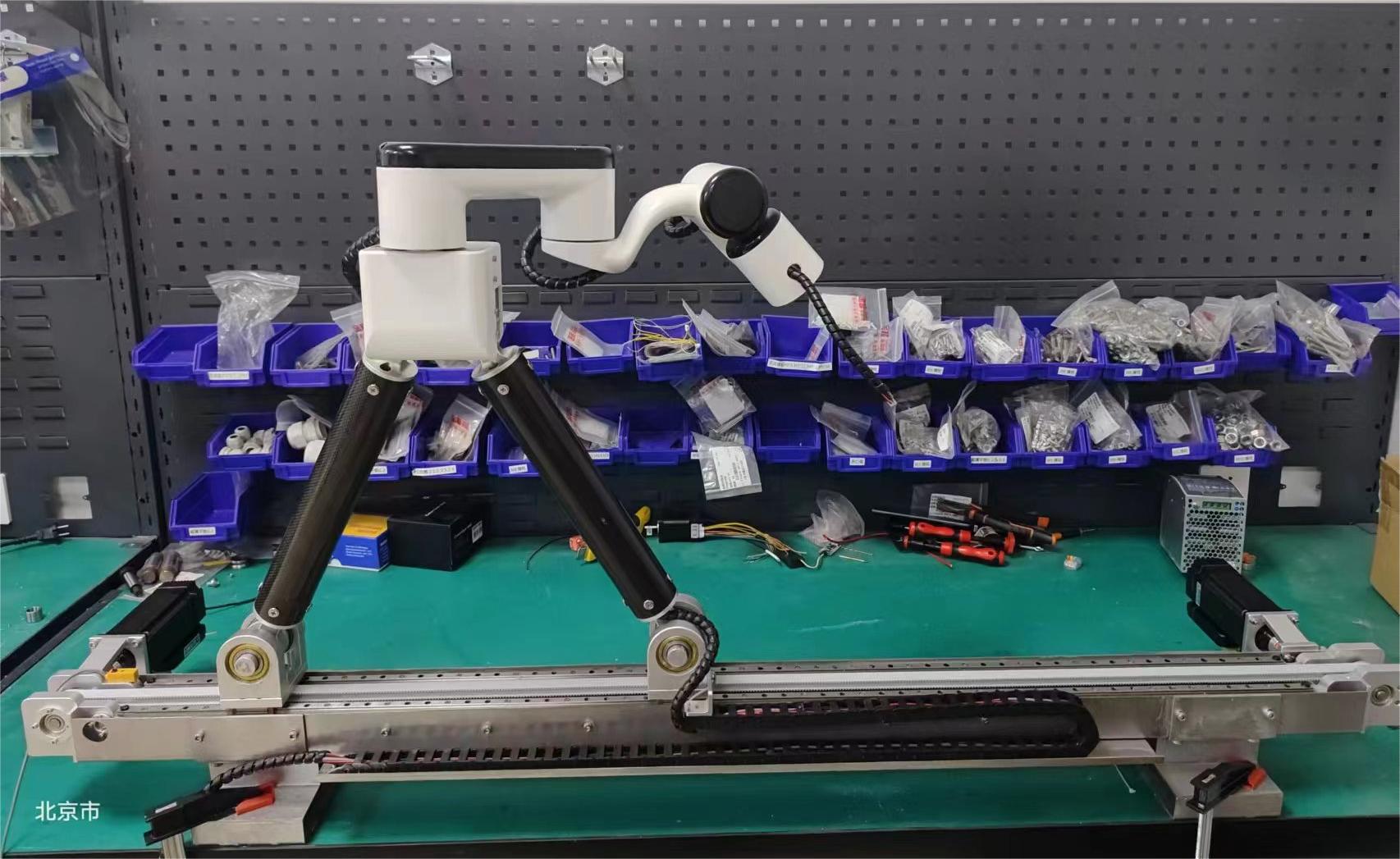}
\caption{The overall assembly of the six-DoF hybrid robotic arm.}
\label{fig6}
\end{figure}
 With an additional two revolute joints, a six-DoF hybrid robotic arm was obtained, as shown in Fig. \ref{fig6}. 
 In the manufacturing process, we utilized lightweight and high-strength materials including aluminum alloy tubes and carbon fiber rods. To ensure high accuracy, stainless steel was used to make the linear rail via the sheet metal process. The black covers on the serial joints were 3D printed. 
 The transmission section incorporated components such as timing belts, gears, incomplete gear sets, and bearings. The electrical system employed a 48V DC stabilized power supply to activate the servo motors, with two DC-DC power adaptors converting voltage from 48V to 24V and 12V for other modules, respectively. 
 Finally, a CAN-USB module facilitated the communication between the robotic arms and a computer for testing purposes.

\section{Kinematic analysis and control of robotic arm}

Prior to developing a control system, it is necessary to model the robot, define the coordinate system for each joint, and then depict the motion relationship among them through coordinate transformation.

\begin{figure}[htbp]
\centering
\includegraphics[scale=0.49]{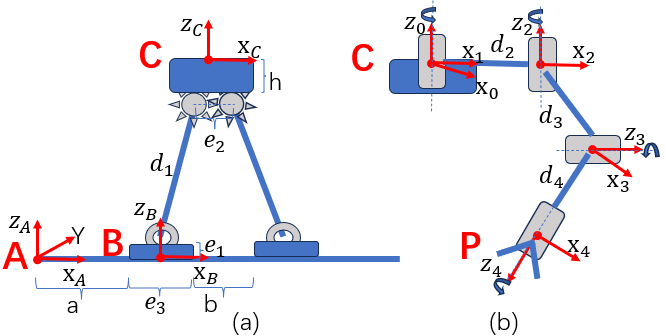}
\caption{Coordinate system of the hybrid robotic arm: a) parallel mechanism; b) serial mechanism.}
\label{fig7}
\end{figure}

\subsection{Kinematic analysis} 

In this study, a simplified geometric model and coordinate system for the robotic arm were established, as shown in Fig. \ref{fig7}. This coordinate system allowed for easy representation and calculation of the geometry and kinematics of the robotic arm using Denavit-Hartenberg (DH) parameters. The arm was segmented into serial and parallel mechanisms, which were independently modeled. 

The upper part of the parallel mechanism moves only within the $x-z$ plane. Consequently, we simplified the parallel component of the dynamic coordinate system \{{B\}-dynamic coordinate system \{C\}\ and selected a fixed coordinate system \{A\} at one end of the guide rail as the reference point, as shown in Fig. \ref{fig8}(a). In this coordinate system, the coordinate values in each direction for point C are shown in formulas (1)–(3):
\begin{equation}
X c=e_3/2+a+(b+e_3-e_2) / 2
\end{equation}

\begin{equation}
Z_c=\sqrt{d_1^2-((b+e_3-e_2) / 2)^2+h+e_1}
\end{equation}

\begin{equation}
Y_c=e_4
\end{equation}
where $e_4$ is the offset of point C with respect to point A in the $y$ axis direction.

The pose matrix established by the DH parameters is shown in  Eq. ( \ref{eq4} ):

\begin{equation}\label{eq4}
_A^CT=\left[\begin{array}{cccc}
1 & 0 & 0 & X_c\\
0 & 1 & 0 & Y_c \\
0 & 0 & 1 &  Z_c \\
0 & 0 & 0 & 1
\end{array}\right]
\end{equation}

Next, we conducted a kinematic analysis of the serial mechanism. As shown in Fig. \ref{fig7}(b), point C was the reference point of the coordinate system. In this coordinate system, the design results using DH parameters are shown in Table \ref{tab_1}.
\begin{table}[H]
\caption{\textbf{DH parameters of serial mechanism.}}
\label{tab_1}
\resizebox{\linewidth}{!}
{
\begin{tabular*}{\linewidth}{@{}ccccc@{}}
\toprule
Link&angle $\theta_i$&Link twist $\alpha_{i-1}$&Link length $a_{i-1}$&Link offset $d_i$\\
\midrule
1&$\theta_1$&0&0&0 \\
2&$\theta_2$&0&$d_2$&0 \\ 
3&$\theta_3$&$90^{\circ}$&$d_3$&0 \\
4&$\theta_4$&$90^{\circ}$&0&$d_4$ \\
\bottomrule
\end{tabular*}
}
\end{table}

The homogeneous transformation matrix from coordinate system \{i-1\} to coordinate system \{i\} was defined as $_{i-1}^{i}$T, as shown in  Eq. ( \ref{eq1} ):

\begin{equation}\label{eq1}
{ }_{i-1}^i T=\left[\begin{array}{cccc}
\cos \theta_i & -\sin \theta_i \cos \alpha_i & \sin \theta_i \sin \alpha_i & a_i \cos \theta_i \\
\sin \theta_i & \cos \theta_i \cos \alpha_i & -\cos \theta_i \sin \alpha_i & a_i \sin \theta_i \\
0 & \sin \alpha_i & \cos \alpha_i & d_i \\
0 & 0 & 0 & 1
\end{array}\right]
\end{equation}

For the convenience of expression, $ci$ represents $\cos \theta_i$, and $si$ represents $\sin \theta_i$. According to (5), formulas (6)-(9) can be derived:

\begin{equation}
_0^1T=\left[\begin{array}{cccc}
c_1 & -s_1 & 0 & 0 \\
s_1 & c_1 & 0 & 0 \\
0 & 0 & 1 &  0 \\
0 & 0 & 0 & 1
\end{array}\right]
\end{equation}

\begin{equation}
_1^2T=\left[\begin{array}{cccc}
c_2 & -s_2 & 0 & c_2d_2 \\
s_2 & c_2 & 0 & s_2d_2 \\
0 & 0 & 1 &  0 \\
0 & 0 & 0 & 1
\end{array}\right]
\end{equation}

\begin{equation}
_2^3T=\left[\begin{array}{cccc}
c_3 & 0 & s_3 & c_3d_3 \\
s_3 & 0 & -c_3 & s_3d_3 \\
0 & 1 & 0 &  0 \\
0 & 0 & 0 & 1
\end{array}\right]
\end{equation}

\begin{equation}
_3^4T=\left[\begin{array}{cccc}
c_4 & 0 & s_4 & 0 \\
s_4 & 0 & -c_4 & 0 \\
0 & 1 & 0 &  d_4 \\
0 & 0 & 0 & 1
\end{array}\right]
\end{equation} 

Then, we multiplied the formulas (2)–(5) to obtain the pose matrix of the end point p with respect to the base point C, as shown in  Eq. ( \ref{eq10} ):
\begin{equation}\label{eq10}
_C^PT=_0^1T * _1^2T * _2^3T * _3^4T
\end{equation}

Based on the analysis above, we can obtain two attitude matrices $_C^PT$ and $_A^CT$. The transformation from the static coordinate system \{A\} through the dual-slider reference coordinate system \{C\} to the end coordinate system \{P\} was achieved by combining the two matrices, as shown in  Eq. ( \ref{eq11} ):
\begin{equation}\label{eq11}
_A^PT=_A^CT * _C^PT=\left[\begin{array}{cccc}
n x & o x & a x & p x \\
n y & o y & a y & p y \\
n z & o z & a z & p z \\
0 & 0 & 0 & 1
\end{array}\right]
\end{equation}

Inverse kinematics is the foundation for motion control and trajectory planning of robotic arms. Since a single position may have multiple valid joint variables, inverse kinematics can be challenging. Two commonly used methods for solving inverse kinematics are analytical and numerical solutions, which are generalized but slow. Given the unique structure of our robotic arm, we investigated the analytical method for solving inverse kinematics. This approach significantly improved the computation speed.

\begin{figure}[htbp]
\centering
\includegraphics[scale=0.38]{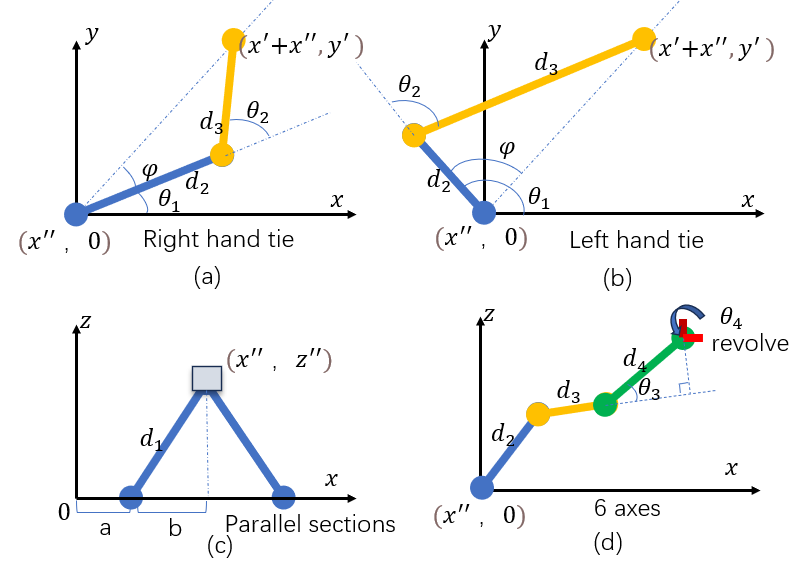}
\caption{Diagram of the proposed solution for inverse kinematics: a) the right-hand system built a model of the serial part; b) the left-hand system built a model of the serial part; c) simplified model of the parallel part; d) simplified model of an improved 6-axis robotic arm.}
\label{fig8}
\end{figure}

Fig. \ref{fig8} illustrates the decomposition of the robotic arm into a serial part and a parallel part. The motion direction of the robotic arm motion was determined in a right-handed and left-handed system, respectively, specified to be positive in the counterclockwise direction.

(Step 1) We solved the serial part. The joint angles were determined by analyzing the position of the robotic arm's end with respect to the coordinate origin using trigonometric functions, as shown in formulas (12)–(14):

\begin{equation}
\theta_1= \begin{cases}\operatorname{atan} \frac{y^{\prime}}{x^{\prime}}-\varphi & \theta_2\geq0 \\ \operatorname{atan} \frac{y^{\prime}}{x^{\prime}}+\varphi 
 &\theta_2<0\end{cases}
\end{equation}

\begin{equation}
c_2=\frac{x^{\prime^2}+y^{\prime^2}-d_2^2-\left(d_3+d_4\cos\theta_3\right)^2}{2 d_2 d_3}
\end{equation}

\begin{equation}
\cos \varphi=\frac{\left(d_3+d_4\cos\theta_3\right)^2-\left(x^{\prime^2}+y^{\prime^2}\right)-d_2^2}{-2 d_2 \sqrt{x^{\prime 2}+y^{\prime 2}}} 
\end{equation}

(Step 2) We solved the parallel part, since it was only in the x-z plane, we followed the simple geometric relationship to obtain the expressions of the joint variables a and b, as shown in  Eq. ( \ref{eq15} ):

\begin{equation}\label{eq15}
\left\{\begin{array}{l}
\sqrt{d_1^2-\left(\frac{b+e_3-e_2}{2}\right)^2}+h+e_1=z^{\prime \prime}=z^{\prime}-d4*\cos\theta_3\\
e_3/2+a+\frac{b+e_3-e_2}{2}+e_2=x^{\prime \prime}
\end{array}\right. 
\end{equation}

(Step 3) According to the above steps, when the pose of the arm was known, the joint variables of the manipulator joints 1 to 6 can be obtained by combining the formulas (12–15). During the solution process, we considered the possibility of multiple valid solutions for the same end posture. To address this, we used joint variable a as the reference variable and traversed its range of values from 0 to 800 mm at 1 mm intervals. We then outputted the sequence of valid solutions for the joint variables, which could be used for subsequent changes in the end pose.

\subsection{Kinematic simulation}
\begin{figure}[htbp]
\centering
\includegraphics[scale=0.62]{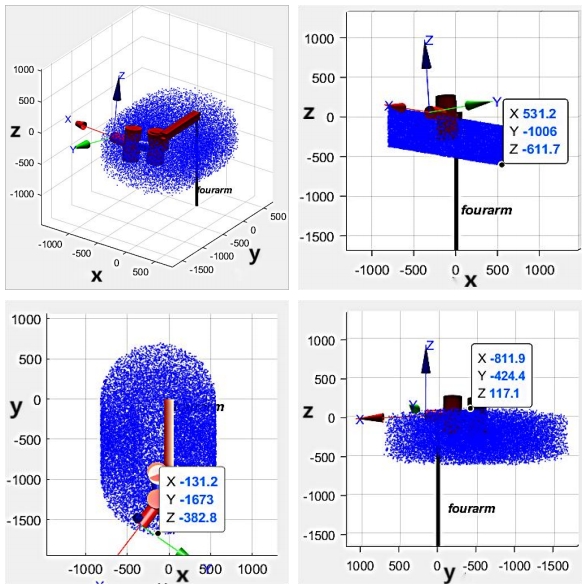}
\caption{The generated point cloud for visualizing the workspace of the hybrid arm.}
\label{fig9}
\end{figure}

To visualize the workspace of the robotic arm, we used a discrete Monte Carlo method combined with inverse kinematics of the mechanism for the hybrid arm's workspace analysis. A set of 1000 randomly generated points was used to show the workspace, as shown in Fig. \ref{fig9}. 

Fig. \ref{fig9} shows that the hybrid mechanism's workspace is irregularly shaped in three dimensions. In the $xoz$ view, the entire structure appears as an approximate ellipse. The longer axis represents the sum of the translation in the direction of the $x$ axis and the radius of the two rotary joints. In the $yoz$ view, it is evident that the working height of the end in the $x-z$ plane varies, allowing easy switching of the working range based on different scenarios and the symmetric arrangement of the arms for collaborative operation. By selecting critical points and coordinate axes in several drawings, we can obtain the range of the three-dimensional envelope space: 
$z\in$[-611.7 mm, 117.1 mm], $x\in$[-812 mm, 535 mm], $y\in$[-1673 mm, 600 mm]. 
This range validated the features of the extensive workspace as designed by the proposed hybrid mechanism.

\begin{figure}[htpb]
\centering
\includegraphics[scale=0.18]{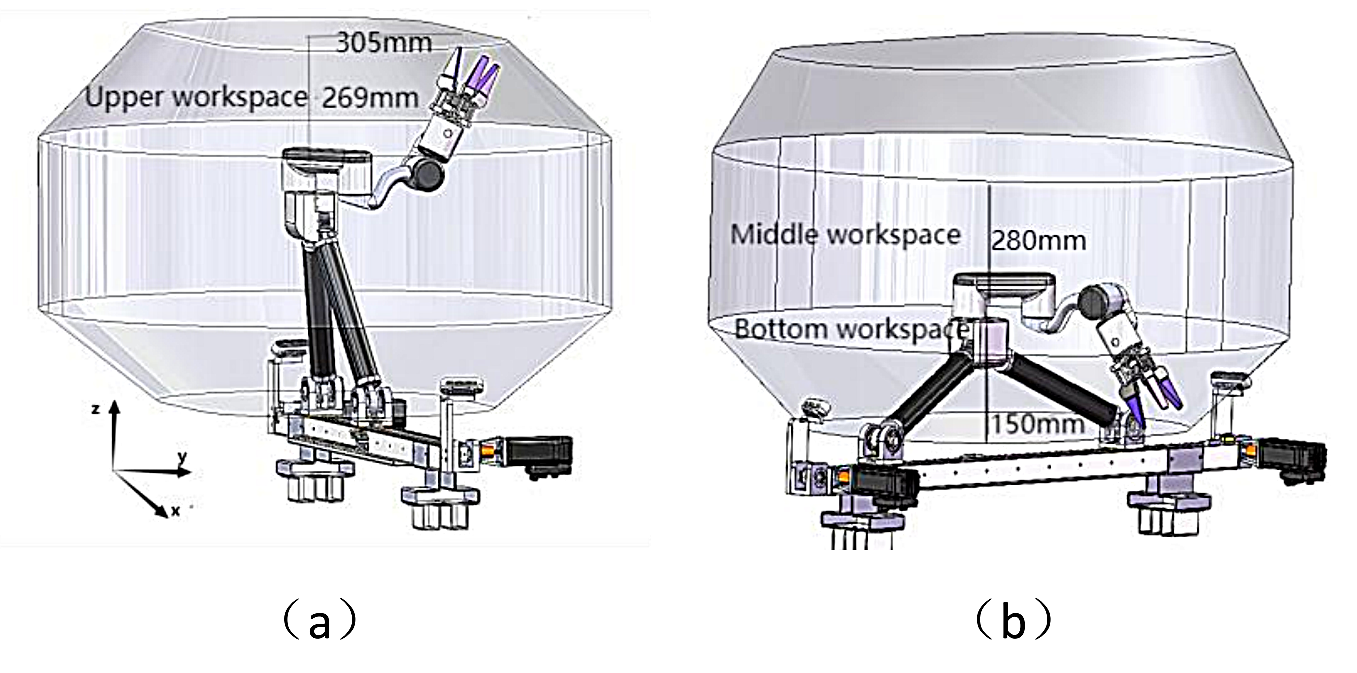}
      \caption{Visualization of the hybrid robotic arm's workspace: a) upper workspace; b) middle and bottom workspace.}
      \label{fig10}
   \end{figure}

Furthermore, we performed a visualization process using modeling software, as depicted in Fig. \ref{fig10}. 
The link length $d_1$, $d_2$, $d_3$ and $d_4$ of our robotic arm were 393 mm, 160 mm, 145 mm and 118 mm, respectively, and the guide rail was up to 1212 mm. 
The overall operational space has an irregular ellipsoidal shape and contains no unreachable holes, unlike many other serial arms. This is because the arm can move freely along the $x$ axis and can access any unreachable point by repositioning itself in the $x$ direction.
In Fig. \ref{fig10}(a), the robotic arm was shown in the upper operating space with a vertical motion range of 269 mm and a minimum operating radius of 305 mm. Fig. \ref{fig10}(b) illustrated the robotic arm in the central operating space with a vertical motion range of 280 mm and a minimum operating radius of 574 mm. 

\begin{table}[H]
\caption{\textbf{Comparison of technical parameters of robotic arms}}
\centering
\label{tab_2}
\begin{tabular}{p{3.1cm}ccc}
\toprule
Brand&Ours&UR3&IRB 360 1-800 \\
\midrule
DoF&6&6&6 \\
Maximum horizontal working radius (mm)&574&500&480 \\
Maximum vertical working
cross-section (mm$^{2}$)&6x10$^{5}$&4x10$^{5}$&1x10$^{5}$ \\
Workspace volume (mm$^{3}$)&1.4x10$^{9}$&5x10$^{8}$&1x10$^{8}$ \\

\bottomrule
\end{tabular}
\end{table}

Compared to similarly sized robotic arms such as the serial UR3, which has a spherical workspace, and the parallel IRB, which has a spherical crown workspace, our arm had an overwhelming advantage in terms of workspace volume, which is almost three times the volume of UR3 serial arms and fourteen times that of IRB parallel arms, as shown in Table \ref{tab_2}.

\subsection{Control system}
\begin{figure}[htbp]
\centering
\includegraphics[scale=0.4]{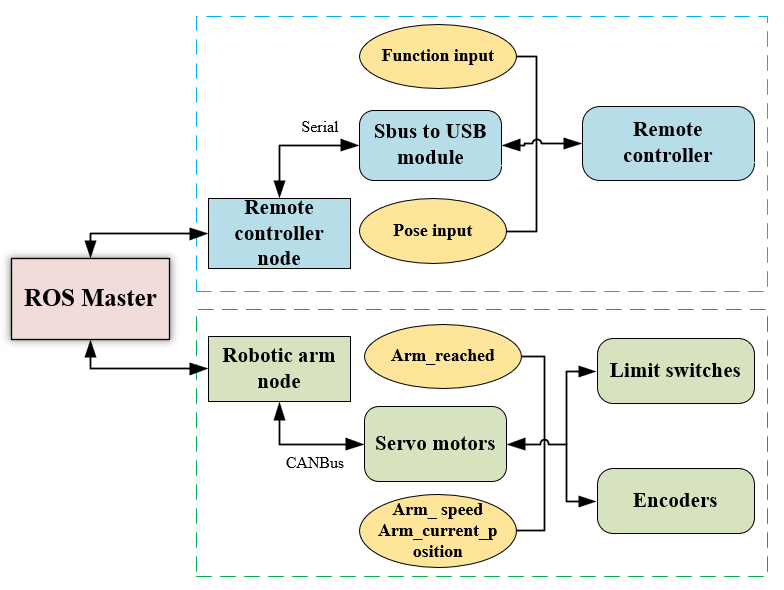}
\caption{Hardware and software architecture of the robotic arm.}
\label{fig11}
\end{figure}

Our application required the hardware communication bus to have a transmission time of under 10 microseconds. The total system response and processing time should not exceed 100 milliseconds, with a maximum delay of 10 milliseconds in generating the target. As shown in Fig. \ref{fig11}, we chose to use Robot Operating System (ROS) to configure the hardware interface of the robotic arm and create corresponding nodes. Sensor data was transmitted to the control node through the controller area network (CAN) bus and serial communication. The core node was responsible for managing the interaction flow of the motors and the remote controller information, including remote control functions, arm position, velocity and other related aspects.

\begin{figure}[htbp]
\centering
\includegraphics[scale=0.34]{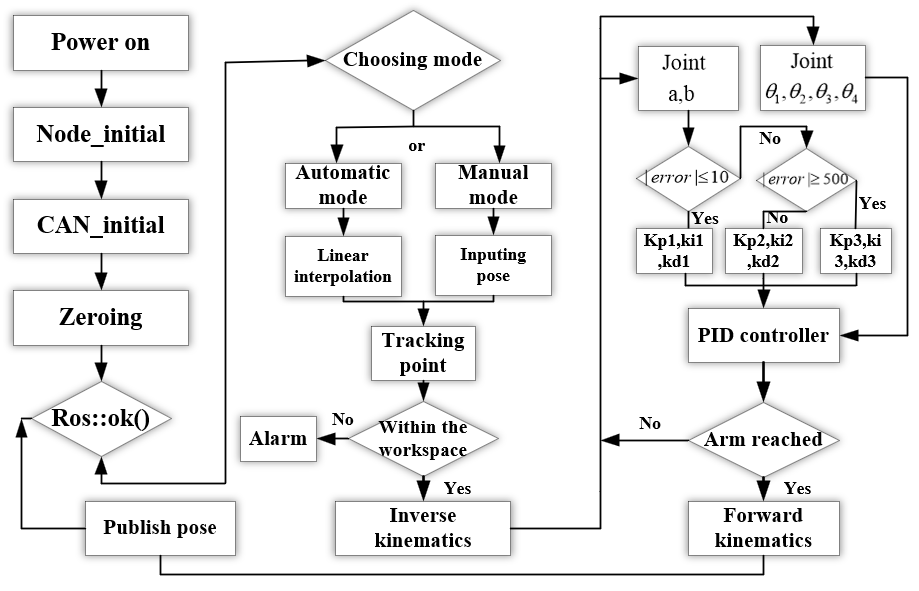}
\caption{Flowchart showing the robotic arm control system.}
\label{fig12}
\end{figure}

Fig. \ref{fig12}  shows the control flow of the robotic arm. 
Operational modes were selected based on the users' requirements, and trajectory points were generated using interpolation algorithms. The inverse kinematics algorithm then converted the desired end-effector positions or orientations into joint angles. Segmented
PID control in the joint space effectively addressed issues of balancing fast motion and quick docking of the sliders caused by the long rail. The guide rail distance was divided into three ranges: [0, 10 mm], [10 mm, 500 mm] and [500 mm, 1212 mm], with corresponding kp, kd, and ki set for each segment. By comparing the actual and expected outputs and adjusting the control error, precise control can be achieved. The joint motors utilized an S-curve acceleration and deceleration algorithm to make the movement smoother. Additionally, real-time feedback control was employed to ensure optimal performance. Position information was captured and sent to the forward kinematic function, allowing for monitoring and adjustment of the robotic arm's motion trajectory within the ROS interface. The arm control loop updated at a frequency of 60 hertz per second.

\section{Experiments and results}


We carried out a series of experiments to evaluate the effectiveness of our robotic arm, including: 1) functional tests to confirm the robotic arm's ability to rotate around a fixed point, duck under obstacles, and maintain a pose when moving with a mobile platform; 2) repeatability tests on each axis and at the end of the robotic arm to show the precision; and 3) trajectory tracking tests to analyze the dynamic stability of the robotic arm during operation.

\subsection{Functional test}

Several initial experiments were carried out to confirm the unique features of the hybrid robotic arm design, including the following tests.

\subsubsection{Rotating around a point}
\begin{figure}[htbp]
\centering
\includegraphics[scale=0.34]{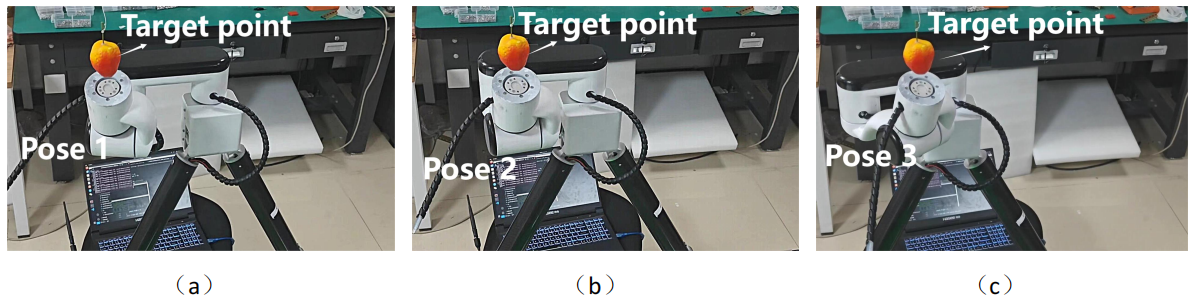}
\caption{The process of the robotic arm rotating around a fixed point: a) the robotic arm was adjusted to orientation 1; b) the robotic arm was adjusted to orientation 2; c) the robotic arm was adjusted to orientation 3.}
\label{fig13}
\end{figure}

To demonstrate the movement of the robotic arm around a fixed point, we placed a strawberry above the robotic arm as the picking target. The operator controlled the robotic arm to reach the designated picking position with the arm's end, making necessary adjustments using orientation transformation functions. The importance of this process is highlighted in Fig. \ref{fig13}, where although the end coordinate position remained constant, there was a noticeable shift in the end's orientations. This adjustment was needed in finding the optimal picking pose.

\subsubsection{Ducking under obstacles}

\begin{figure}[htbp]
\centering
\includegraphics[scale=0.21]{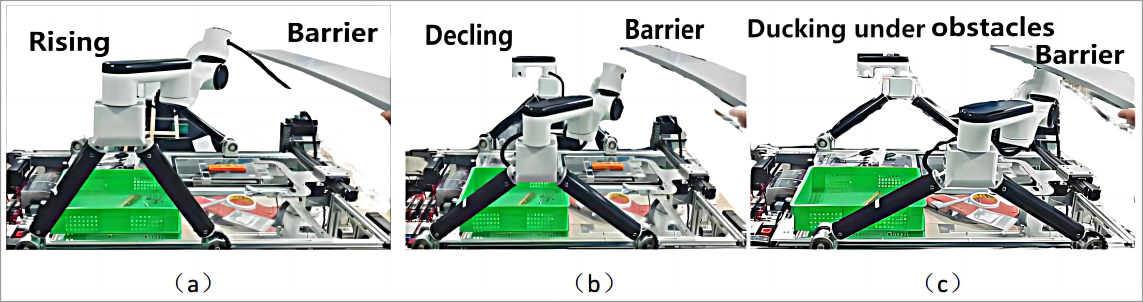}
\caption{Ducking under obstacles: a) the robotic arm stands at a normal working height; b) doing a split to lower itself to an appropriate height; c) passing under the table along the rail.}
\label{fig14}
\end{figure}
To demonstrate the obstacle avoidance capability of the robotic arm, we conducted a simulation of strawberry growing environment in the laboratory. Using a horizontal bar to represent the strawberry table, Fig. \ref{fig14} demonstrates how the robotic arm ducks under obstacles. The robotic arm began at a normal working height and then smoothly performed a split to lower itself, ducking under the table as it moved along the rail. This demonstration highlighted the impressive obstacle avoidance ability of the robotic arm, a feat not achievable by other serial or parallel arms.

\subsubsection{Moving with pose unchanged}
\begin{figure}[htbp]
\centering
\includegraphics[scale=0.36]{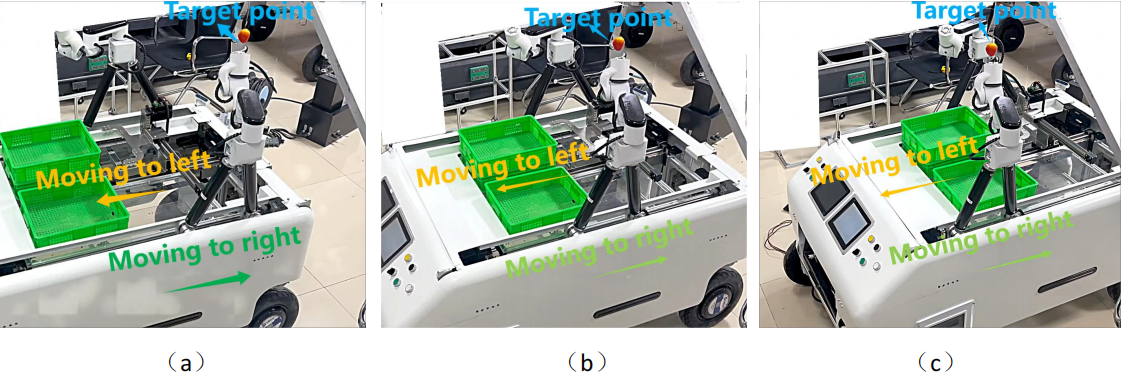}
\caption{The robotic arm keeping its end pose unchanged while moving with a mobile base: a) the robotic arm is picking a target with an initial end pose; b) the mobile base moves forward while the robotic arm moves to the left; c) the arm's end pose remains unchanged.
}
\label{fig15}
\end{figure}

To demonstrate the robotic arm's ability of keeping its end pose unchanged while moving with a mobile base, we mounted it on a laboratory-developed mobile base and placed a strawberry above it as the target for picking. As shown in Fig. \ref{fig15}, the robotic arm was controlled to reach the picking point with an initial end pose, then the mobile base moved to the right, while the robotic arm simultaneously moved to the left. During this process, the end of the robotic arm remained pose unchanged at the picking point. This unique feature enabled the robotic arm to stay at a point and keep an optimal orientation for picking for a while even when the chassis is continuously moving forward.

 It should be noted that the robotic arm vibrated slightly when the trajectory was moving, which was mainly due to the closed-loop position control and the disturbance force of the rigid structure. 
 
\begin{figure}[htbp]
\centering
\includegraphics[scale=0.12]{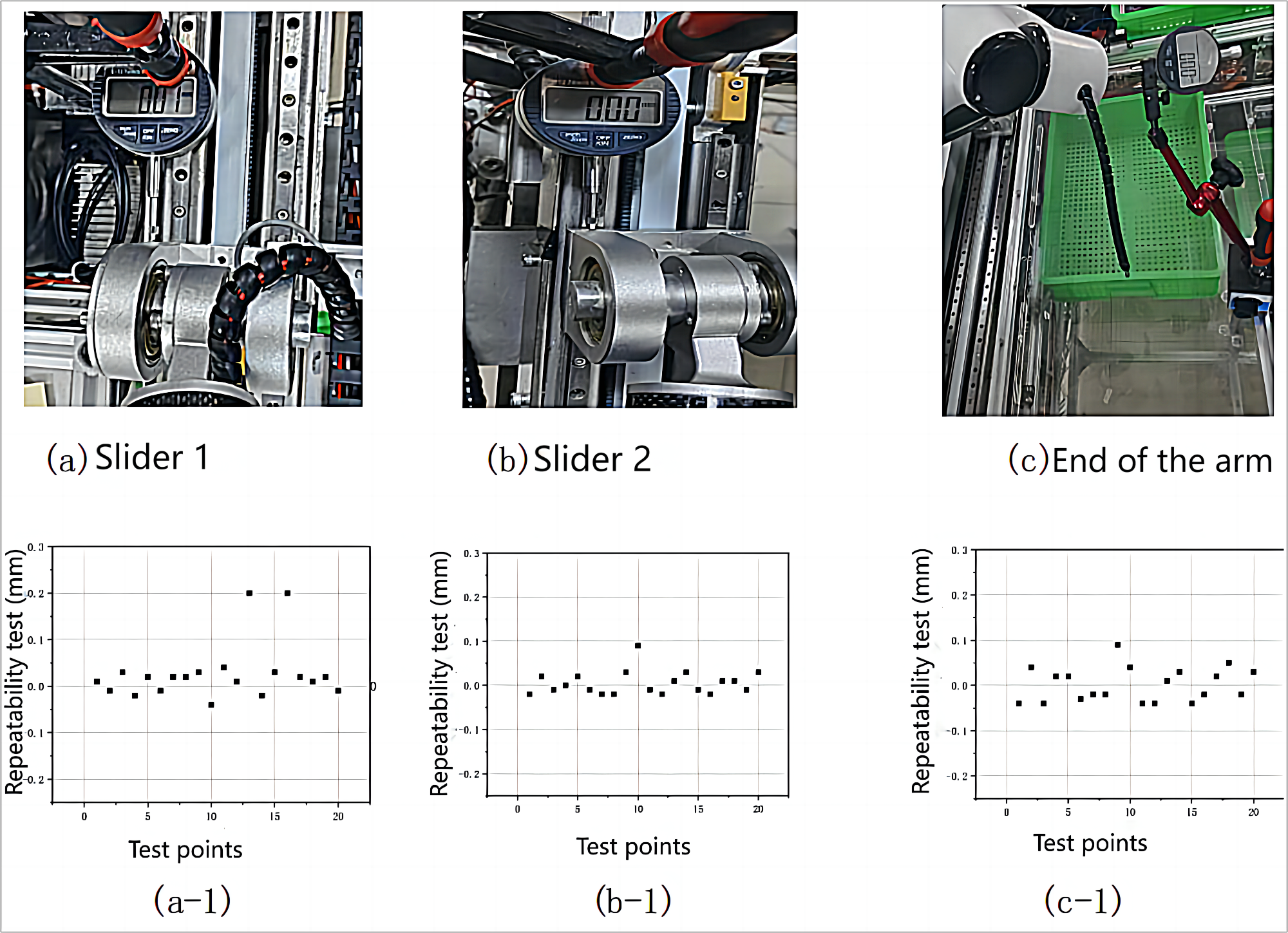}
\caption{Repeatability test.}
\label{fig16}
\end{figure}
\subsection{Repeatability test}

To evaluate the positioning accuracy of the robotic arm, a repeatability test was performed on the robotic arm system, which independently tested two horizontal axes and the end of the arm. As shown in Fig. \ref{fig16}, a dial indicator was attached to the arm, and when in motion, the shafts touch the indicator's tip. Before starting the test, we zeroed the motors of each joint in the robotic arm. 
The graph shows the results of the repeatability test after zero-means normalization for 20 trials at a maximum running speed of 150 mm/s. The repeatability of slider 1, slider 2 and the arm's end were 0.017 mm, 0.03 mm and 0.109 mm, respectively. The difference in shaft-to-shaft accuracy was mainly due to the various transmission types and ratios, but all of these accuracies are sufficient for our harvesting applications.

\subsection{Trajectory tracking test}

We conducted a point-to-point positioning test on the robotic arm system to evaluate its dynamic performance. By specifying target positions, we directed the robotic arm to these locations and recorded the actual trajectories. As illustrated in Fig. \ref{fig17}, during the experiment, we commanded the arm to move from coordinate point A (300 mm, 200 mm, 400 mm) to B (580 mm, 200 mm, 400 mm). The trajectory planning process employed linear interpolation algorithms, generating 10 trajectory points to run sequentially. Real-time tracking of the actual position of the robotic arm was achieved through sensors, facilitating the comparison between the actual and expected positions.

\begin{figure}[htbp]
\centering
\includegraphics[scale=0.1]{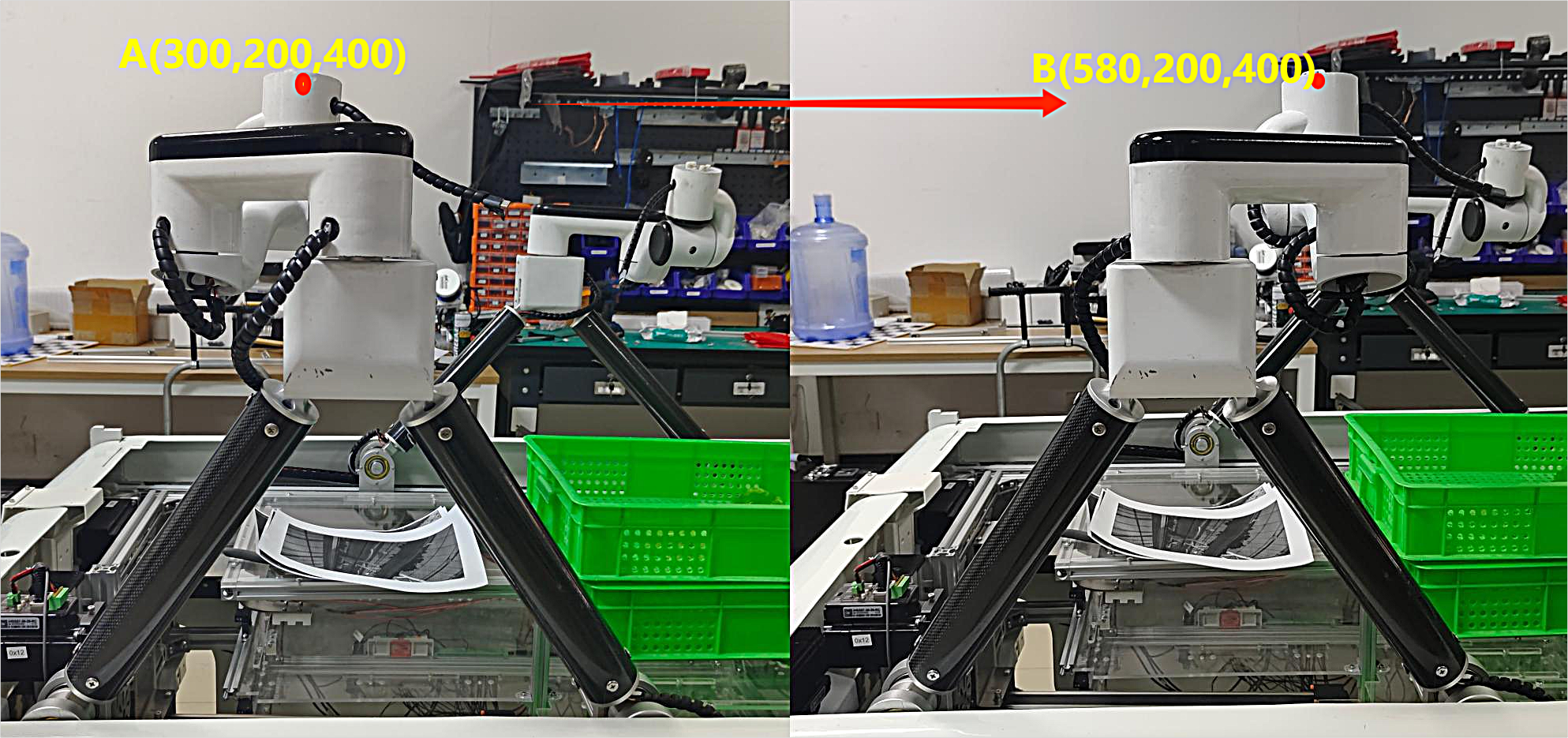}
\caption{The robotic arm moves linearly from point A to point B.}
\label{fig17}
\end{figure}

Fig. \ref{fig18}(a) demonstrates the rapid response curves of the two sliders of the single-rail platform 
using the segmented PID control algorithm. Both sliders can rapidly respond and stabilize after reaching the designated target positions. Fig. \ref{fig18}(b) illustrates the variation in angular velocity of the upper revolute joint with respect to angular displacement. Compared to the set target speed, a smoother curve was observed, indicating stable acceleration and deceleration of the rotary motor, thereby avoiding end-effector jitter. In Fig. \ref{fig18}(c), the blue curve represents the planned linear trajectory, while the red curve represents the actual trajectory measured during the trajectory tracking process. Minor deviations occurred due to the acceleration and deceleration of the motors and the root mean squared error
(RMSE) calculated from the trajectory was approximately 0.38 mm, which was sufficient for obstacle avoidance and path planning during our harvesting process.

\begin{figure}[htbp]
\centering

\includegraphics[scale=0.25]{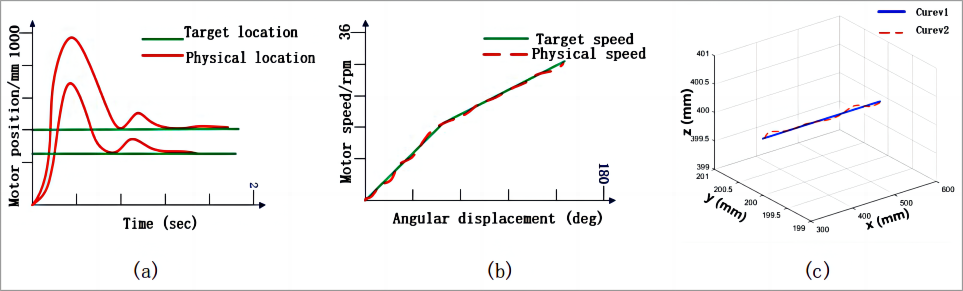}
\caption{The speed and position feedback during the operation of the arm: a) the speed curves of the two sliders using the segmented PID control algorithm; b) the rotational speed of the revolute joint with respect to the angular displacement; c) trajectory tracking during the linear movement.}
\label{fig18}
\end{figure}

Through the aforementioned experiments, we obtained the performance parameters of the robotic arm, as shown in Table \ref{tab_3} below, including the positioning accuracy of each axis, maximum speed, operational workspace, and others.
\begin{table}[H]
\caption{\textbf{Technical parameters of six-DoF robotic arm}}
\centering
\label{tab_3}
\begin{tabular}{ccc}
\toprule
parameter&value&unit \\
\midrule
Slider 1 repetition precision&0.017&mm \\ 
Slider 2 repetition precision&0.03&mm \\
End repetition precision&0.109&mm \\
Maximum horizontal working radius&574&mm \\
Maximum vertical working cross-section&1212x500&mm$^{2}$ \\
Workspace volume &1.4x10$^{9}$&mm$^{3}$ \\
Speed&1.5&m/s \\
Payload&3&kg \\
Average power&680&w \\

\bottomrule
\end{tabular}
\end{table}

\section{conclusion}

A novel hybrid six-DoF robotic arm was proposed that merged the benefits of both parallel and serial mechanisms. A meshed-gear parallel mechanism was designed capable of lowering itself and performing a split for passing under obstructions, such as pipes, tables and beams typically found in greenhouses. The parallel mechanism, coupled with linkages, used a meshed-gear set to maintain the upper part parallel to the rail, offering a compact alternative to traditional large parallel linkage mechanisms. Moreover, through the integration of serially mounted joints, the hybrid arm was able to retain the end's pose even when it moved with a mobile platform, making it possible for fruit picking using the optimal pose for a while in dynamic conditions. Also, the hybrid arm's workspace appeared like an irregular ellipsoidal shape, being almost three times the workspace volume of similar sized UR3 serial arms and fourteen times that of the ABB IRB parallel arms. Experiments also showed the repeatability errors were 0.017 mm, 0.03 mm and 0.109 mm for the two sliders and the arm's end, respectively, demonstrating the arm's high precision for agricultural applications.

Future efforts will involve integrating sensors around the arm and creating a reinforcement learning-based algorithm to improve its ability to avoid obstacles.

\addtolength{\textheight}{-12cm}   







\bibliographystyle{unsrt}
\bibliography{sample}
\end{document}